\documentclass[final]{colt2023} 

\usepackage{nicefrac}
\usepackage{times}

\renewcommand{\le}{\leqslant}
\renewcommand{\leq}{\leqslant}

\renewcommand{\geq}{\geqslant}

\newcommand{\qed}{\hfill$\blacksquare$}
\newcommand{\eps}{\varepsilon}

\newcommand{\E}{\mathbb E}
\newcommand{\R}{\mathbb R}

\newcommand{\Z}{\mathsf Z}

\newcommand{\B}{\mathcal B}
\newcommand{\F}{\mathcal F}
\newcommand{\N}{\mathcal N}

\newcommand{\W}{\mathcal W}

\newcommand{\A}{\mathsf A}
\renewcommand{\H}{\mathsf H}

\newcommand\dd{{\rm d}}

\def\argmin{\operatornamewithlimits{argmin}}
\def\esssup{\operatornamewithlimits{esssup}}

\newtheorem{Th}{Theorem}[section]
\newtheorem{Lem}[Th]{Lemma}

\newtheorem{Prop}[Th]{Proposition}

\newtheorem{Co}[Th]{Corollary}

\newtheorem{Ex}[Th]{Example}
\newtheorem{As}[Th]{Assumption}

\title[Exploring Local Norms in Exp-concave Statistical Learning]
{Exploring Local Norms in Exp-concave Statistical Learning}

\author{%
    \Name{Nikita Puchkin} \Email{npuchkin@hse.ru}\\
    \addr HSE University and IITP RAS, Russian Federation%
    \AND
    \Name{Nikita Zhivotovskiy} \Email{zhivotovskiy@berkeley.edu}\\
    \addr UC Berkeley, Department of Statistics, USA%
}

\begin{document}

\maketitle

\begin{abstract}%
We consider the problem of stochastic convex optimization with exp-concave losses using Empirical Risk Minimization in a convex class. Answering a question raised in several prior works, we provide a
\[
O\left(\frac{d}{n} + \frac{1}{n}\log\left(\frac{1}{\delta}\right)\right)
\]
excess risk bound valid for a wide class of bounded exp-concave losses, where $d$ is the dimension of the convex reference set, $n$ is the sample size, and $\delta$ is the confidence level. Our result is based on a unified geometric assumption on the gradient of losses and the notion of local norms. 
\end{abstract}


\section{Problem statement}

We consider a stochastic convex optimization problem, that is, minimization of the objective
\[
    F(w) = P f(w, Z)
\]
over a compact \emph{convex} set $\W \subset \R^d$. Here $Z$ is a random element with a distribution $P$ supported on a set $\Z$, and $P f(w, Z)$ stands for the expectation of $f(w, Z)$ with respect to $Z \sim P$. Since $P$ is usually unknown, the optimizer
\begin{equation}
    \label{eq:w_star}
    w^* \in \argmin\limits_{w \in \W} F(w)
\end{equation}
cannot be computed explicitly. Instead, a learner has an access to a sample $S_n = \{ Z_i : 1 \leq i \leq n \}$ with independent elements distributed according to $P$. The most natural and popular strategy in this case is to replace the expectation with respect to $Z \sim P$ by the sample mean and study an empirical risk minimizer (ERM)
\begin{equation}
\label{eq:erm}
    \widehat w \in \argmin\limits_{w \in \W} \widehat F(w),
\end{equation}
where
\begin{equation}
\label{eq:widehatf}
    \widehat F(w)
    = P_n f(w, Z)
    = \frac1n \sum\limits_{i=1}^n f(w, Z_i)
    \quad \text{for all $w \in \W$.}
\end{equation}
Here and further in the paper, $P_n$ stands for the empirical measure, corresponding to the sample $S_n$. We are interested in tight high probability bounds on the excess risk. Our main result will imply a bound of the form 
\[
F(\widehat w) - F(w^*) = O\left(\frac{d}{n} + \frac{1}{n}\log\left(\frac{1}{\delta}\right)\right),
\]
for a general class of bounded exp-concave losses.

\paragraph{Contributions.}
Our contribution combines two ideas: specific assumptions on the loss function and a careful probabilistic analysis utilizing empirical process techniques. We propose a general assumption on loss functions that applies to most commonly used exp-concave losses in the bounded setup. 
Our main result, Theorem \ref{th:excess_risk}, addresses the question  raised explicitly in a series of papers \citep*{mahdavi2015lower, koren15, mehta17, gonen18, yang2018simple} by further exploring the concept of local norms. Despite the presence of established empirical process techniques in related contexts, our approach handles the loss functions exhibiting data-dependent curvature. Compared to the existing localized analysis for the losses with (data-independent) curvature \citep{mendelson2002geometric,sridharan08,koltchinskii2011oracle}, we do not assume any linear structure, so the common approach based on computing the local complexity of the linear class is not applicable in our situation. Similarly, some more recent applications of empirical process techniques to exp-concave losses \citep{vijaykumar21} do not fully exploit the local geometric structure of the reference set and thus lead to suboptimal bounds in our setup. We discuss this in more detail in Section \ref{sec:proof}.

\paragraph{Notation.}
Throughout the paper, the notation $f 
\lesssim g$ or $g \gtrsim f$ means that for some universal constant $c > 0$ we have $f \leq cg$. For any set $\W \subseteq \mathbb{R}^d$, $\textrm{Int}(\W)$ denotes its interior. For a symmetric positive semidefinite matrix $\A \in \R^{d \times d}$, $\|\cdot\|_\A$ stands for a seminorm, induced by $\A$:
\[
    \|w\|_\A = \sqrt{w^\top \A w},
    \quad \text{for all $w \in \W$}.
\]
Finally, for any two positive semidefinite matrices $\A$ and $\mathsf{B}$ of the same size the inequalities $\A \preceq \mathsf{B}$ and $\mathsf{B} \succeq \A$ mean that $(\mathsf{B} - \A)$ is positive semidefinite.

\paragraph{Structure of the paper.} The rest of the paper is organized as follows. In Section \ref{sec:relliterature}, we make a brief overview of related works. In Section \ref{sec:learning}, we introduce our geometric assumption, present our main result, Theorem \ref{th:excess_risk}, and discuss its implications. Section \ref{sec:proof} is devoted to the proof of Theorem \ref{th:excess_risk}. It heavily relies on a technical statement about the supremum of a certain empirical process, Lemma \ref{th:exp_moment}. We provide its proof in Section \ref{sec:exp_moment_proof}. Proofs of auxiliary results appear in Appendix.

\section{Related work}
\label{sec:relliterature}
Theoretical properties of the empirical risk minimizer $\widehat w$, especially its generalization ability, were extensively studied in the contexts of stochastic convex optimization and statistical learning theory. Given the extensive body of relevant literature, we review only a subset of the most relevant papers. Early results in this direction include the analysis of linear regression with squared loss via local Rademacher complexities (see, e.g., \citep{mendelson2002geometric} and \citep[Section 5.1]{koltchinskii2011oracle}).
In particular, \citet[Section 5.1, Example 1]{koltchinskii2011oracle} shows that, in the problem of bounded linear regression in $\mathbb{R}^d$, the excess squared risk of a ERM scales as $O(d / n)$ with high probability. In contrast to the standard results on learning general VC-type classes under various margin conditions (see, e.g., \citep*{bartlett05, massart06, zhivotovskiy18}, and \citep[Section 5.1, Example 3]{koltchinskii2011oracle}), the linear regression specific rate $O(d / n)$  avoids an extra $\log n$ factor and exhibits tighter bounds on generalization ability of ERM. The reason for that is a linear structure of the reference set, which allows to bound its local Rademacher complexity using the standard Cauchy-Schwarz or Fenchel inequalities (see \citep[Proposition 3.2]{koltchinskii2011oracle} and \citep[Theorem 1]{kakade08b}). 

Many researchers have further explored the possibility of achieving an excess risk bound better than $O(d \log n / n)$ for problems with specific local structures in the reference set. Recent investigations in stochastic approximation lead to numerous tight in-expectation $O(d/n)$ bounds on the excess risk of particular optimization algorithms (see, e.g., \citep{moulines11, bach2013non, woodworth21} to mention a few). The current paper focuses on high probability bounds, which typically require a more technical analysis.
In a study by \citet*{sridharan08}, the authors extended the analysis of local Rademacher complexities to the case of regularized strongly convex losses, $f(w, z) = \ell(y, w^\top x) + R(w)$, where $z = (x, y)$, $\ell$ is convex with respect to its second argument for any $y$, and $R(w)$ is strongly convex. Similar to an example in \citep[Section 5.1, Example 1]{koltchinskii2011oracle}, the key to the proof was a bound on the local Rademacher complexity of a linear class, as demonstrated in Lemma 7 in \citep{sridharan08}. The PAC-Bayesian localization technique was introduced by \citet{audibert2010linear, audibert2011robust} to obtain high-probability bounds on the squared excess risk of ERM in the bounded linear regression problem. These authors achieved a rate of $O((d + \log(1/\delta)) / n)$. This approach has also been used in \citep{zhang2006information} in the context of density estimation. Related to our study, \citet{klochkov21} proved a dimension-free excess risk bound of $O(\log(n) \log(1/\delta)/n)$ with high probability for arbitrary strongly convex and Lipschitz losses, using the uniform stability of ERM. We also would like to mention that excess risk bounds can be derived from regret bounds for online learning algorithms via the online-to-batch conversion. We refer to \citep{kakade08, gaillard18, van2022regret} for related high probability upper bounds. However, in our context, the online-to-batch approach yields risk bounds with additional logarithmic factors, which makes it unsuitable for our purposes.

This paper focuses on stochastic convex optimization problems using a broad class of exp-concave losses. Our goal is to provide tight high probability bounds $O((d + \log(1/\delta)) / n)$ on the excess risk of ERM under mild assumptions on the function $f(w, z)$. In the context of exp-concave losses, this question was directly addressed in a series of papers and turned out to be quite challenging. 
It appears that one also needs to make some additional assumptions on the loss function that allows to capture the local geometry of the reference set.  Several works have achieved the desired $O(d/n)$ rate in expectation (e.g., \citep{koren15, gonen18, zhang19}), but transforming this guarantee into a high probability bound results in suboptimal $O((d \log n + \log(1/\delta)) / n)$ guarantees. Several groups of authors \citep*{mahdavi2015lower, koren15, mehta17, gonen18, yang2018simple} have questioned the possibility of achieving the optimal $O((d + \log(1/\delta)) / n)$ high probability excess risk bound. The best bound in this direction is given by the confidence boosting approach suggested by \cite{mehta17}, which replaces ERM with an aggregate of independently trained ERMs, resulting in a bound of $O(d\log(1/\delta) / n)$. We additionally refer to the recent papers by \cite*{mourtada2022improper} and \cite*{bilodeau2021minimax} where a detailed survey of relevant results is provided. 

The cornerstone of the analysis of empirical risk minimizers in learning problems with the squared loss (or more general strongly convex losses) is localization. Certain fixed point equations appeared in influential works \citep{van1990estimating, birge1993rates, yang1999information} and then became familiar to the machine learning community thanks to the papers \citep*{koltchinskii2006local, bartlett05}. We refer to the monographs \citep{van2000empirical, koltchinskii2011oracle} and the paper \citep{mendelson2002geometric} for examples of application of local entropy and local Rademacher complexity to the analysis of linear regression with squared loss. 
Finally, we note that the analysis of the local structure of the reference set in the context of density estimation has its roots in the classical works of \cite{lecam1973convergence} and \cite{birge1983approximation}. For related upper bounds, we refer to Section 32.2.3 in the forthcoming book by \citet{polyanskiy2023}.

\section{Main results and examples}
\label{sec:learning}

In this section, we present our main result.
Throughout the paper, we assume that the loss function $f(w, z)$ satisfies the following condition.

\begin{As}
    \label{as:sc}
    For any training sample $Z_1, \dots, Z_n$ there is positive semidefinite matrix $\H \in \R^{d \times d}$, which may depend on the sample, such that 
    \begin{itemize}
        \item the expectation of $\H$ is finite;
        \item $\widehat F(w)$ given by \eqref{eq:widehatf} is $\sigma$-strongly convex on $\W$ with respect to the seminorm, induced by $\H$; that is, for any $u, v \in \W$ and $\alpha \in [0, 1]$, the next inequality holds almost surely:
        \begin{equation}
            \label{eq:sc}
            \widehat F \big(\alpha u + (1 - \alpha) v \big)
            \leq \alpha \widehat F(u) + (1 - \alpha) \widehat F(v) - \frac{\sigma \alpha (1 - \alpha)}2 \|u - v\|_\H^2;
        \end{equation}
        \item for any $u, v \in \W$, the empirical $L_2$-distance between $f(u, Z)$ and $f(v, Z)$ satisfies the inequality
        \[
            P_n \big( f(u, Z) - f(v, Z) \big)^2
            \leq L^2 \|u - v\|_\H^2.
        \]
    \end{itemize}
    Here, $\sigma$ and $L$ are positive constants that do not depend on $Z_1, \dots, Z_n$.
\end{As}
When $\H$ is the identity matrix, we simply say that $\widehat F$ is a strongly convex function. Similarly, when $\H$ is the identity matrix, the second part of this assumption corresponds to the standard Lipschitz property.
Though Assumption \ref{as:sc} may appear similar to the commonly used assumptions of strong convexity and Lipschitzness, the introduction of the data-dependent matrix $\H$ broadens the range of applications of our model. In particular, it includes regularized objectives, considered in \citep{sridharan08}, and generalized linear models (see, e.g., \citep{gonen18}), where the loss function has a form
\begin{equation}
    \label{eq:glm_loss}
    f(w, z) = \ell(w^\top x, y).
\end{equation}
Here $z = (x, y)$ is a feature-label pair, $w$ and $x$ belong to bounded subsets of $\R^d$, and $\ell$ is $L$-Lipschitz and $\sigma$-strongly convex with respect to the first argument for all $y$. In this case, $\widehat F(w)$ is $\sigma$-strongly convex with respect to the seminorm $\|\cdot\|_\H$, induced by
$\H = (X_1 X_1^\top + \dots + X_n X_n^\top) / n$ (see Example \ref{ex:logreg}).
Our setup covers both of the most commonly used generalized linear models: linear regression, which corresponds to $f(w, z) = (y - w^\top x)^2$, and logistic regression, which corresponds to $f(w, z) = \log(1 + e^{-y w^\top x})$ with $y \in \{-1, 1\}$. 

Under mild differentiability assumptions, $\widehat{F}$ demonstrates exponential concavity, as shown in the following result.

\begin{Prop}
    \label{prof:exp-concavity}
    Suppose that $\widehat F$ satisfies Assumption \ref{as:sc} and let the loss function $f(w, z)$ be twice differentiable with respect to the first argument for all $z \in \Z$. Then $\widehat F$ is $(\sigma / L^2)$-exponentially concave on $\W$. That is, $e^{-\sigma \widehat F / L^2}$ is concave on $\W$.
\end{Prop}
However, our assumptions are slightly stronger than exp-concavity alone. In fact, our assumption implies the boundedness of the loss.
\begin{Prop}
    \label{prop:boundedness}
    Under Assumption \ref{as:sc}, we have $|f(w, Z) - f(w^*, Z)| \leq 4 L^2 / \sigma$ for all $w \in \W$ and almost surely with respect to $Z$.
\end{Prop}
Observe that this result does not necessarily imply the boundedness of the set $\mathcal W$. This distinguishes our assumptions from the more restrictive strongly convex Lipshitz assumption, which implies both the boundedness of the loss and $\mathcal W$. In fact,  the proof of Proposition \ref{prop:boundedness} yields that
\[
    \|w - \widehat w\|_{\H}^2
    \leq \frac{4}{\sigma} \left( \widehat F(w) - \widehat F(\widehat w) \right)
    \leq \frac{32 L^2}{\sigma^2}
    \quad \text{for all $w \in \W$}.
\]
If $\det(\H) \neq 0$, then $\W$ is bounded. However, in general, the matrix $\H$ is not assumed to be nondegenerate. Hence, $\W$ may potentially be unbounded under our assumption.  The proofs of Propositions \ref{prop:boundedness} and \ref{prof:exp-concavity} are deferred to Appendix \ref{app:propositions}. After listing these basic properties, we are ready to present our main result.
\begin{Th}
    \label{th:excess_risk}
    Assume that $\W \subset \R^d$ is a compact convex set.
    Under Assumption \ref{as:sc}, for any $\delta \in (0, 1)$, it holds that, with probability at least $1 - \delta$, the ERM estimator \eqref{eq:erm} satisfies
    \[
        F(\widehat w) - \min\limits_{w \in \mathcal W}F(w) \lesssim \frac{L^2 (d + \log(1/\delta))}{\sigma n}.
    \]
\end{Th}

Theorem \ref{th:excess_risk} answers a natural question raised in several previous papers. The most relevant to our result are the papers of \citet{koren15, gonen18}, where the authors obtained optimal in-expectation bounds of order $O(d/n)$, and of \citet{mahdavi2015lower, mehta17, yang2018simple}, where the authors proved $O((d\log n + \log(1/\delta))/n)$ high probability upper bounds. We will now discuss the differences in assumptions made on the loss functions.
In \citep{koren15, yang2018simple}, the authors required the gradient $\partial f(w, z) / \partial w$ be Lipschitz on $\W$. \cite{koren15} also asked if the smoothness of the gradient is necessary for obtaining fast rates for ERM. A partial answer was given in \citep{gonen18}, where the authors considered generalized linear models with the losses of the form \eqref{eq:glm_loss}, where $\ell(\cdot, y)$ is Lipschitz and strongly convex with respect to its first argument for all $y$'s. Under these assumptions, the authors showed that the expected excess risk of ERM can decay as fast as $O(d/n)$. Our result strictly improves the one of \cite{gonen18}. As it was discussed earlier, our Assumption \ref{as:sc} is more general than the requirements of \cite{gonen18}. At the same time, we derive a rate optimal $O((d + \log(1/\delta))/n)$ high probability excess risk bound. Concerning the papers \citep{mahdavi2015lower, mehta17}, the authors work under more general assumptions than Assumption \ref{as:sc}. Namely, they only assume that the loss $f(w, z)$ is bounded, Lipschitz and $\sigma$-exponentially concave with respect to $w \in \W$ for all $z \in \Z$. It is not clear, if it is possible to get rid of extra $\log n$ factors in their high probability bounds. However, we would like to note that our setup covers all interesting bounded loss functions arising in exp-concave learning with convex reference sets. 

Theorem \ref{th:excess_risk} has several implications. We discuss them in the examples below.

\begin{Ex}[Linear regression with squared loss] 
    Fix $R > 0$.
    Assume that the observations $Z_1, \dots, Z_n$ have a form $Z_i = (X_i, Y_i)$, $1 \leq i \leq n$, where $X_i \in \R^d$ and $Y_i \in \R$ have distributions supported on $\{x \in \R^d : \|x\| \leq 1\}$ and $[-R, R]$, respectively. Consider $\W = \{w \in \R^d : \|w\| \leq R\}$ and the quadratic loss function $f(w, z) = (y - w^\top x)^2$. Note that $f(w, z)$ fulfils Assumption \ref{as:sc} with the matrix $\H = (X_1 X_1^\top + \dots + X_n X_n^\top) / n$
    and the constants $\sigma = 2$, $L = 4R$. According to Theorem \ref{th:excess_risk}, for any $\delta \in (0, 1)$, an empirical risk minimizer $\widehat w$ satisfies the inequality
    \[
        F(\widehat w) - F(w^*) \lesssim \frac{R^2(d + \log(1/\delta)}n
    \]
    with probability at least $1 - \delta$.
    This agrees with the upper bound of \citet[Section 5.1, Example 1]{koltchinskii2011oracle}, based on local Rademacher complexities, and the lower bound of \citet[Theorem 1]{shamir15}, who showed that for any estimator $\widetilde w \in \mathcal W$ there exists a data distribution over $\B(0, 1) \times [-R, R]$ such that
    $\E F(\widetilde w) - F(w^*) \gtrsim R^2 d / n$.
\end{Ex}

\begin{Ex}[Logistic regression]
\label{ex:logreg}
    Again, assume that the observations $Z_1, \dots, Z_n$ have a form $Z_i = (X_i, Y_i)$, $1 \leq i \leq n$, where $X_i \in \R^d$ and $Y_i \in \{-1, 1\}$. Fix $R > 0$ and let the distribution of $X_i$'s be supported on $\B(0, R) = \{x \in \R^d : \|x\| \leq R\}$. Consider $\W = \{w \in \R^d : \|w\| \leq B\}$ and the logistic loss
    \[
        f(w, z) = \log(1 + e^{-y w^\top x}).
    \]
    Direct calculations show that the map $u \mapsto \log(1 + e^u)$ is $1$-Lipschitz and $e^{-BR}$-strongly convex on $[-BR, BR]$. Hence, the logistic loss meets Assumption \ref{as:sc} with the matrix
    \[
        \H = \frac1n \sum\limits_{i = 1}^n X_i X_i^\top
    \]
    and the constants $\sigma = e^{-BR}$, $L = 1$. Theorem \ref{th:excess_risk} implies that, for any $\delta \in (0, 1)$, with probability at least $1 - \delta$, it holds that
    \begin{equation}
    \label{eq:logisticregression}
        F(\widehat w) - F(w^*) \lesssim \frac{e^{BR} (d + \log(1/\delta))}n.
    \end{equation}
    We refer to \citep*{hazan14}, where the authors discuss the inavoidability of the $e^{BR}$ factor for ERM. We note that the bound \eqref{eq:logisticregression} has not appeared in the literature before.
\end{Ex}

Furthermore, our bound covers strongly convex and Lipschitz losses, regardless of any linear structure in the loss function. In \citep{klochkov21}, the authors used the uniform stability argument to prove that, if $f(\cdot, z)$ is $L$-Lipschitz and $\sigma$-strongly convex for all $z \in \Z$, then, for any $\delta \in (0, 1)$, with probability at least $1 - \delta$, it holds that
\begin{equation}
\label{eq:klzh}
    F(\widehat w) - F(w^*) \lesssim \frac{L^2 \log n\log(1 / \delta)}{\sigma n}.
\end{equation}
Taking together their result and Theorem \ref{th:excess_risk}, we get the following bound.

\begin{Co}[Strongly convex and Lipschitz losses]
    \label{co:strongly_convex}
    Assume that, for any $z \in \Z$, the loss function $f(\cdot, z)$ is $L$-Lipschitz and $\sigma$-strongly convex on $\W \subseteq \mathbb{R}^d$, which is a compact convex set. Then, for any $\delta \in (0, 1)$, with probability at least $1 - \delta$, it holds that
    \[
        F(\widehat w) - F(w^*) \lesssim \frac{L^2\min\{d + \log(1/\delta), \; \log n\log(1/\delta)\} }{\sigma n}.
    \]
\end{Co}
This gives the best known high probability upper bound in this standard setup. Recent results on algorithmic stability \citep{bousquet2020sharper, klochkov21} leave the question if the $\log n$ term can be removed in the bounds such as \eqref{eq:klzh}. Our Corollary \eqref{co:strongly_convex} shows that if this factor cannot be removed, then any demonstration of this must occur in the regime $d + \log(1/\delta) \gtrsim \log n\log(1/\delta)$.

\paragraph{On logarithmic factors and improper learning.}
We focus on the sharp analysis of ERM in a convex reference set $\mathcal W$. Nevertheless, recent work has demonstrated that the use of so-called \emph{improper} estimators --- estimators that predict values outside of the reference set $\mathcal W$ --- can lead to significant improvements in the dependence on some parameters. These estimators are typically based on aggregating the functions within certain \emph{non-convex} reference sets constructed by truncating the original functions that correspond to $\mathcal W$. Such improvements are known in both logistic regression \citep{kakade2004online, foster2018logistic, mourtada2022improper} and regression with the squared loss \citep{forster2002relative, vavskevivcius2023suboptimality, Mourtada2021}. However, it should be noted that the improper learning setup poses a considerable challenge for the localized analysis utilized in this paper. Specifically, the lower bound of \citet*[Theorem 6]{rakhlin2017empirical} indicates that, in the context of bounded regression with a non-convex reference set, the lower bound of $\Omega(\log (n)/n)$ cannot be surpassed even with the use of improper learners. At present, there is no known analysis that can yield high probability excess risk bounds of order $O((d + \log(1/\delta))/n)$ for any improper learner in non-convex reference sets.

\section{Proof of Theorem \ref{th:excess_risk}}
\label{sec:proof}

Our proof is making most of the standard arguments on the Laplace transform of \emph{shifted} or \emph{offset} empirical processes. A similar approach was exploited by many authors \cite*{wegkamp2003model, lecue2014optimal, liang15, zhivotovskiy18, kanade2022exponential} in the statistical setup, though their analysis is specific to strongly convex losses or binary losses under additional probabilistic assumptions. In \citep{vijaykumar21}, the author generalized the approach of \cite{liang15} to study the properties of ERM under exp-concave losses, though their analysis only focuses on getting the $O(1/n)$ rate of convergence and fails to capture the local structure of the reference set. In our context their result would give an additional $\log n$ factor, which is known to be achievable using an $\varepsilon$-net argument \citep{mehta17}. The key point of our analysis is based on relating Assumption \ref{as:sc} with the localized covering numbers of the set $\mathcal W$ with respect to the data-dependent semi-norm induced by $\H$.  Two papers that are worth mentioning on the topic of online learning and exp-concave losses, which utilize sequential offset processes, are \citep{rakhlin2015sequential} and \citep{bilodeau2020tight}.

\begin{Lem}
    \label{lem:offset_symmetrization}
    Under Assumption \ref{as:sc}, 
    let $\Phi : \R \rightarrow \R$ be a convex monotonously increasing function. Then, it holds that
    \[
        \E \Phi \big(F(\widehat w) - F(w^*) \big)
        \leq \E \E_\eps \Phi \left( 4 \sup\limits_{w \in \W} \left[ P_n \eps \big( f(w, Z) - f(w^*, Z) \big) - \frac{\sigma}8 \|w - w^*\|_\H^2 \right] \right),
    \]
    where $\eps_1, \dots, \eps_n$ are i.i.d. Rademacher random variables.
\end{Lem}
The proof of Lemma \ref{lem:offset_symmetrization} is postponed to Appendix \ref{app:offset_symmetrization_proof}.
From now on, we focus our attention on the shifted multiplier process of the form
$P_n \eps \big( f(w, Z) - f(w^*, Z) \big) - \sigma \|w - w^*\|_\H^2 / 8, w \in \W$,
and consider exponential moments of its supremum. In Section \ref{sec:exp_moment_proof}, we prove our main technical bound.

\begin{Lem}
    \label{th:exp_moment}
    Let $f(w, z)$ satisfy Assumption \ref{as:sc}.
    Set $\lambda = \sigma n / (32 e L^2)$.
    Then, for all $Z_1, \dots, Z_n$, it holds that
    \[
        \E_\eps \exp\left\{ \lambda \sup\limits_{w \in \W} \left[ P_n \eps \big( f(w, Z) - f(w^*, Z) \big) - \frac{\sigma}8 \|w - w^*\|_\H^2 \right] \right\}
        \leq e + e^{3 d} + 14 e^{2048 (1 + e)^2 d / e}.
    \]
\end{Lem}

Let us briefly describe the idea of the proof of Lemma \ref{th:exp_moment}. We use the standard peeling argument and represent the whole set $\W$ as the union
\[
    \W = \W[0, r] \cup \left( \bigcup\limits_{k = 0}^\infty \W[2^k r, 2^{k+1} r] \right),
\]
where $r > 0$ is a fixed number and, for any $b \geq a \geq 0$, $\W[a, b]$ is defined as
\[
    \W[a, b] = \left\{ w \in \W : a \leq \|w - w^*\|_\H \leq b \right\}. 
\]
After this step, we essentially reduce the problem to the analysis of localized offset process, that is, the process of the form
\[
    P_n \eps \big( f(w, Z) - f(w^*, Z) \big) - \frac{\sigma}8 \|w - w^*\|_\H^2,
\]
where $w$ runs over a local set $\W[0, r]$, rather than over the whole set $\W$. However, the peeling argument alone is not sufficient to avoid unnecessary $\log n$ factors, and one should be careful at this point. In \cite[Theorem 1]{mehta17} and \cite[Lemma 6, Lemma 7]{liang15}, the authors use the $\eps$-net argument, which leads to a suboptimal result in our situation. The main difference of our approach from the previous works is hidden in the proof Lemma \ref{lem:local_chaining}, where we combine the localized analysis with chaining and Talagrand's inequality to derive a tight upper bound on the exponential moment
\[
    \E_\eps \exp\left\{ \lambda \sup\limits_{w \in \W[0,r]} P_n \eps \big( f(w, Z) - f(w^*, Z) \big) \right\}.
\]
Given that we consider the geometry of the loss in defining $\W[0, r]$, deriving an upper bound on its covering number is relatively straightforward. We provide this bound in Lemma \ref{lem:covering}. The key point of this approach, which sets it apart and enables improvement over existing results, is the decision not to use the naive $\eps$-net argument.

Lemma \ref{lem:offset_symmetrization} and Lemma \ref{th:exp_moment} immediately imply that
\[
    \log \E \exp \left\{\lambda (F(\widehat w) - F(w^*)) \right\} \lesssim d
    \quad \text{with $\lambda = \frac{\sigma n}{128 e L^2}$.}
\]
Hence, for any $\delta \in (0, 1)$, the following inequality holds, with probability at least $1 - \delta$:
\[
    F(\widehat w) - F(w^*)
    \lesssim \frac{L^2 (d + \log(1/\delta))}{\sigma n}.
\]
The claim follows.

\hfill\qed

\section{Proof of Lemma \ref{th:exp_moment}}
\label{sec:exp_moment_proof}

Our proof relies on the classical peeling argument. For any $b \geq a \geq 0$ define
\[
    \W[a, b] = \left\{ w \in \W : a \leq \|w - w^*\|_\H \leq b \right\}. 
\]
Let $r > 0$ be a positive real number to be specified later. Since the function $\Phi(x) = e^{\lambda x}$ takes only positive values, the expectation of interest does not exceed
\begin{align}
    \label{eq:exp_moment_sum}
    &\notag
    \E_\eps \exp\left\{ \lambda \sup\limits_{w \in \W} \left[ P_n \eps \big( f(w, Z) - f(w^*, Z) \big) - \frac{\sigma}8 \|w - w^*\|_\H^2 \right] \right\}
    \\&
    \leq \E_\eps \exp\left\{ \lambda \sup\limits_{w \in \W[0, r]} \left[ P_n \eps \big( f(w, Z) - f(w^*, Z) \big) - \frac{\sigma}8 \|w - w^*\|_\H^2 \right] \right\}
    \\&\notag\quad
    + \sum\limits_{k = 0}^\infty \E_\eps \exp\left\{ \lambda \sup\limits_{w \in \W[2^k r, 2^{k+1} r]} \left[ P_n \eps \big( f(w, Z) - f(w^*, Z) \big) - \frac{\sigma}8 \|w - w^*\|_\H^2 \right] \right\}.
\end{align}   
Thus, we have bounded the exponential moment of the supremum of the empirical process by the sum of exponential moments of localized processes.
The main ingredient in the proof of Lemma \ref{th:exp_moment} is Lemma \ref{lem:local_chaining}, which provides an upper bound on the exponential moment of the supremum of a localized set without introducing additional logarithmic factors.

\begin{Lem}
    \label{lem:local_chaining}
    Let $f(w, Z)$ satisfy Assumption \ref{as:sc}. Let
    \[
        B = \sup\limits_{w \in \W[0, r]} \esssup\limits_{z \in \Z} |f(w, z) - f(w^*, z)|.
    \]
    Then, for any $\lambda \geq 0$, it holds that
    \begin{align*}
        &
        \log \E_\eps \exp\left\{ \lambda \sup\limits_{w \in \W[0,r]} P_n \eps \big( f(w, Z) - f(w^*, Z) \big) \right\}
        \\&
        \leq 64 \lambda Lr \sqrt{\frac{d}n} + \frac{B^2 \lambda^2 e^{B\lambda/n}}{2n}\left( \frac{128 Lr}{B} \sqrt{ \frac{d}n} + \frac{L^2 r^2}{B^2} \right).
    \end{align*}
    If, in addition, $\lambda \leq n/B$, then
    \[
        \log \E_\eps \exp\left\{ \lambda \sup\limits_{w \in \W[0,r]} P_n \eps \big( f(w, Z) - f(w^*, Z) \big) \right\}
        \leq 64 (1 + e) \lambda Lr \sqrt{\frac{d}n} + \frac{e \lambda^2 L^2 r^2}{2n}.
    \]
\end{Lem}
We defer the proof of Lemma \ref{lem:local_chaining} to Section \ref{sec:local_chaining_proof} and focus on finishing the proof of Lemma \ref{th:exp_moment}. Note that Proposition \ref{prop:boundedness} implies that $B \leq 4 L^2 / \sigma$ under Assumption \ref{as:sc}. Since we consider $\lambda = \sigma n / (32 e L^2)$, the conditions of the second part of Lemma \ref{lem:local_chaining} are satisfied. Applying it to the first term in the right-hand side of \eqref{eq:exp_moment_sum}, we obtain that
\begin{align}
    \label{eq:exp_moment_0r}
    &\notag
    \E_\eps \exp\left\{ \lambda \sup\limits_{w \in \W[0, r]} \left[ P_n \eps \big( f(w, Z) - f(w^*, Z) \big) - \frac{\sigma}8 \|w - w^*\|_\H^2 \right] \right\}
    \\&
    \leq 
    \E_\eps \exp\left\{ \lambda \sup\limits_{w \in \W[0, r]} P_n \eps \big( f(w, Z) - f(w^*, Z) \big) \right\}
    \\&\notag
    \leq \exp\left\{ 64 (1 + e) \lambda Lr \sqrt{\frac{d}n} + \frac{e \lambda^2 L^2 r^2}{2n} \right\}.
\end{align}
Next, fix a non-negative integer $k$ and consider
\[
    \E_\eps \exp\left\{ \lambda \sup\limits_{w \in \W[2^k r, 2^{k+1} r]} \left[ P_n \eps \big( f(w, Z) - f(w^*, Z) \big) - \frac{\sigma}8 \|w - w^*\|_\H^2 \right] \right\}.
\]
In contrast to \eqref{eq:exp_moment_0r}, dropping the negative term leads to a suboptimal bound in this situation. At the same time, the following simple observation allows us to obtain a tight bound.
By the definition of $\W[2^k r, 2^{k+1} r]$, we have $\|w - w^*\|_\H \geq 2^k r$ for any $w \in \W[2^k r, 2^{k+1} r]$. Therefore,
\begin{align*}
    &
    \E_\eps \exp\left\{ \lambda \sup\limits_{w \in \W[2^k r, 2^{k+1} r]} \left[ P_n \eps \big( f(w, Z) - f(w^*, Z) \big) - \frac{\sigma}8 \|w - w^*\|_\H^2 \right] \right\}
    \\&
    \leq \E_\eps \exp\left\{ \lambda \sup\limits_{w \in \W[2^k r, 2^{k+1} r]} \left[ P_n \eps \big( f(w, Z) - f(w^*, Z) \big)\right] - 4^{k-1}\sigma \lambda r^2 / 2\right\}
    \\&
    \leq \E_\eps \exp\left\{ \lambda \sup\limits_{w \in \W[0, 2^{k+1} r]} \left[ P_n \eps \big( f(w, Z) - f(w^*, Z) \big)\right] - 4^{k-1} \sigma \lambda r^2 / 2\right\}.
\end{align*}
Applying Lemma \ref{lem:local_chaining} again, we obtain that
\begin{align}
    \label{eq:exp_moment_k}
    &\notag
    \E_\eps \exp\left\{ \lambda \sup\limits_{w \in \W[0, 2^{k+1} r]} \left[ P_n \eps \big( f(w, Z) - f(w^*, Z) \big)\right] - 4^{k-1} \sigma \lambda r^2 / 2 \right\}
    \\&
    \leq \exp\left\{ 64 (1 + e) \; 2^{k + 1} \lambda Lr \sqrt{\frac{d}n} + \frac{e \cdot 4^{k + 1} \lambda^2 L^2 r^2}{2n} - 4^{k-1} \sigma \lambda r^2 / 2  \right\}
    \\&\notag
    = \exp\left\{ 128 (1 + e) \; 2^{k} \lambda Lr \sqrt{\frac{d}n} + \frac{2e \cdot 4^{k} \lambda^2 L^2 r^2}{n} - 4^{k-1} \sigma \lambda r^2 / 2  \right\}.
\end{align}
Taking \eqref{eq:exp_moment_sum}, \eqref{eq:exp_moment_0r}, and \eqref{eq:exp_moment_k} together, we get that
\begin{align*}
    &
    \E_\eps \exp\left\{ \lambda \sup\limits_{w \in \W} \left[ P_n \eps \big( f(w, Z) - f(w^*, Z) \big) - \frac{\sigma}8 \|w - w^*\|_\H^2 \right] \right\}
    \\&
    \leq \exp\left\{ 64 (1 + e) \lambda Lr \sqrt{\frac{d}n} + \frac{e \lambda^2 L^2 r^2}{2n} \right\}
    \\&\quad
    + \sum\limits_{k=0}^\infty \exp\left\{ 128 (1 + e) \; 2^{k} \lambda Lr \sqrt{\frac{d}n} + \frac{2e \cdot 4^{k} \lambda^2 L^2 r^2}{n} - 4^{k-1} \sigma \lambda r^2 / 2  \right\}.
\end{align*}
The remainder of the proof relies on standard computations and on making the appropriate choice for the parameters $\lambda$ and $r$.
Let us recall that $\lambda = \sigma n / (32 e L^2)$.
It is easy to observe that such a choice ensures
\[
    \frac{2e \cdot 4^k \lambda^2 L^2 r^2}{n}
    = \frac{4^k \sigma \lambda r^2}{16}. 
\]
Then, it holds that
\begin{align*}
    &
    \exp\left\{ 64 (1 + e) \lambda Lr \sqrt{\frac{d}n} + \frac{e \lambda^2 L^2 r^2}{2n} \right\}
    \\&\quad
    + \sum\limits_{k=0}^\infty \exp\left\{ 128 (1 + e) \; 2^k \lambda Lr \sqrt{\frac{d}n} + \frac{2 e \cdot 4^k \lambda^2 L^2 r^2}{n} - 4^{k-1} \sigma \lambda r^2 / 2 \right\}
    \\&
    \leq \exp\left\{ 64 (1 + e) \lambda Lr \sqrt{\frac{d}n} + \frac{\sigma \lambda r^2}{64} \right\}
    + \sum\limits_{k=0}^\infty \exp\left\{ 128 (1 + e) \; 2^k \lambda Lr \sqrt{\frac{d}n} - 4^{k-2} \sigma \lambda r^2 \right\}.
\end{align*}
Choosing
\[
    r = \frac{L}{\sigma} \sqrt{\frac{d}{n}},
\]
we obtain that
\[
    64 (1 + e) \lambda Lr \sqrt{\frac{d}n} + \frac{\sigma \lambda r^2}{64}
    = \frac{64 (1 + e) \lambda L^2 d}{\sigma n} + \frac{\lambda L^2 d}{64 \sigma n}
    = \frac{2 (1 + e) d}{e} + \frac{d}{2048 e}
    < 3 d,
\]
and, for any $k \geq 0$, it holds that
\[
    128 (1 + e) \; 2^k \lambda Lr \sqrt{\frac{d}n} - 4^{k-2} \sigma \lambda r^2
    \leq \frac{\lambda L^2 d}{\sigma n} \left( 128 (1 + e) \; 2^k - 4^{k-2} \right).
\]
Thus, we conclude that
\begin{align}
    \label{eq:exp_moment_bound}
    &\notag
    \E_\eps \exp\left\{ \lambda \sup\limits_{w \in \W} \left[ P_n \eps \big( f(w, Z) - f(w^*, Z) \big) - \frac{\sigma}8 \|w - w^*\|_\H^2 \right] \right\}
    \\&
    \leq e^{3 d} + \sum\limits_{k=0}^\infty \exp\left\{ \frac{\lambda L^2 d}{\sigma n} \left( 128 (1 + e) \; 2^k - 4^{k-2} \right) \right\},
\end{align}
where $\lambda = \sigma n / (32e L^2)$. It remains to bound the sum
\[
    \sum\limits_{k=0}^\infty \exp\left\{ \frac{\lambda L^2 d}{\sigma n} \left( 128 (1 + e) \; 2^k - 4^{k-2} \right) \right\}.
\]
Note that, if $k \geq 14$, then $128 (1 + e) \; 2^k < 2^{k+9} \leq 4^{k - 2} / 2$. Hence,
\begin{align}
    \label{eq:sum_split}
    &\notag
    \sum\limits_{k=0}^\infty  \exp\left\{ \frac{\lambda L^2 d}{\sigma n} \left( 128 (1 + e) \; 2^k - 4^{k-2} \right) \right\}
    \\&
    \leq \sum\limits_{k=0}^{13}  \exp\left\{ \frac{\lambda L^2 d}{\sigma n} \left( 128 (1 + e) \; 2^k - 4^{k-2} \right) \right\} + \sum\limits_{k=14}^\infty \exp\left\{ -\frac{4^k \lambda L^2 d}{32 \sigma n} \right\}.
\end{align}
Applying the standard bound for the quadratic function
\[
    128(1 + e) \; 2^k - 4^{k - 2}
    = \left.\left( 128(1 + e) x - \frac{x^2}{16} \right)\right|_{x = 2^k}
    \leq 256^2 (1 + e)^2,
\]
we deduce that the first term in the right-hand side does not exceed
\begin{equation}
    \label{eq:first_term_bound}
    14 \exp\left\{ \frac{256^2 (1 + e)^2 \lambda L^2 d}{\sigma n} \right\}
    = 14 \exp\left\{ \frac{2048 (1 + e)^2 d}{e} \right\}.
\end{equation}
For the second one, we have
\begin{align}
    \label{eq:second_term_bound}
    \sum\limits_{k=14}^\infty \exp\left\{ -\frac{4^k \lambda L^2 d}{32 \sigma n} \right\}
    &\notag
    =\sum\limits_{k=0}^\infty \exp\left\{ -\frac{4^{k + 14} d}{32^2 e} \right\}
    \leq \sum\limits_{k=0}^\infty \exp\left\{ -\frac{4^{14} \; k d}{32^2 e} \right\} \\&
    \leq \sum\limits_{k=0}^\infty \exp\left\{ -\frac{4^{9} \; k d}{e} \right\} \le e.
\end{align}
The inequalities \eqref{eq:exp_moment_bound}, \eqref{eq:sum_split}, \eqref{eq:first_term_bound}, and \eqref{eq:second_term_bound} imply that
\[
    \E_\eps \exp\left\{ \lambda \sup\limits_{w \in \W} \left[ P_n \eps \big( f(w, Z) - f(w^*, Z) \big) - \frac{\sigma}8 \|w - w^*\|_\H^2 \right] \right\}
    \leq e + e^{3 d} + 14 e^{2048 (1 + e)^2 d / e}.
\]
The claim follows.

\qed

\subsection{Proof of Lemma \ref{lem:local_chaining}}
\label{sec:local_chaining_proof}

We start with an upper bound on the expected value of the supremum. Let $\F[0, r] = \{f(w, z) - f(w^*, z) : w \in \W[0, r]\}$. According to Assumption \ref{as:sc}, we have
\[
    P_n \big( f(u, Z) - f(v, Z) \big)^2
    \leq L^2 \|u - v\|_\H^2
    \quad \text{for all $u, v \in \W$.}
\]
This yields that, for any $\gamma > 0$,
a $(\gamma / L)$-net in $\W[0, r]$ with respect to the seminorm $\|\cdot\|_\H$ induces a $\gamma$-net in $\F[0, r]$ with respect to the empirical $L_2$-norm.
Indeed, let $\W_{\gamma/L}[0, r]$ be a $(\gamma/L)$-net in $\W[0,r]$. Consider any $g \in \F[0, r]$. By the definition of $\F[0, r]$, there exists $w \in \W[0, r]$, such that $g(z) = f(w, z)$. Let $w_{\gamma/L}$ be the closest to $w$ element of $\W_{\gamma/L}[0, r]$. By the definition of $\W_{\gamma/L}[0, r]$, we have 
$
    \|w - w_{\gamma/L}\|_\H \leq \gamma / L.
$
Then, it holds that
\begin{align*}
    &
    \left| \sqrt{P_n \big( f(w, Z) - f(w^*, Z) \big)^2}
    - \sqrt{P_n \big( f(w_{\gamma/L}, Z) - f(w^*, Z) \big)^2} \right|
    \\&
    \leq \sqrt{P_n \big( f(w, Z) - f(w_{\gamma/L}, Z) \big)^2}
    \leq L \|w - w_{\gamma/L}\|_\H
    \leq \gamma.
\end{align*}
Hence, the covering number of $\F[0, r]$ with respect to the empirical $L_2$-norm does not exceed the one of $\W[0, r]$ with respect to the seminorm $\|\cdot\|_\H$.
Applying the standard chaining technique, we obtain that
\begin{equation}
    \label{eq:dudley}
    \E_\eps \sup\limits_{w \in \W[0,r]} P_n \eps \big( f(w, Z) - f(w^*, Z) \big)
    \leq \frac{12}{\sqrt n} \int\limits_{0}^{Lr} \sqrt{ \log \N(\W[0, r], \|\cdot\|_\H, \gamma/L)} \dd \gamma.
\end{equation}
We use the following lemma to bound the covering number of $\W[0, r]$.

\begin{Lem}
    \label{lem:covering}
    For any $u \in (0, r]$, it holds that
    $
        \N(\W[0, r], \|\cdot\|_\H, u) \leq \left( 6r/u \right)^d.
    $
\end{Lem}
The proof of Lemma \ref{lem:covering} is moved to Section \ref{sec:covering_proof}. Substituting the bound of Lemma \ref{lem:covering} into the Dudley integral \eqref{eq:dudley}, we obtain that
\[
    \E_\eps \sup\limits_{w \in \W[0,r]} P_n \eps \big( f(w, Z) - f(w^*, Z) \big)
    \leq \frac{12}{\sqrt n} \int\limits_{0}^{Lr} \sqrt{ d \log (6Lr/\gamma)} \dd \gamma.
\]
Substituting $\gamma$ with $6Lr e^{-u}$, we get
\[
    \int\limits_{0}^{Lr} \sqrt{\log (6Lr/\gamma)} \dd \gamma
    = 6Lr \int\limits_{\log 6}^{+\infty} \sqrt{u} e^{-u} \dd u
    \leq 6Lr \int\limits_{0}^{+\infty} \sqrt{u} e^{-u} \dd u
    = 6Lr \; \Gamma(3/2) = 3Lr \sqrt{\pi}.
\]
Therefore, we have
\begin{equation}
\label{eq:sup_expectation}
    \E_\eps \sup\limits_{w \in \W[0,r]} P_n \eps \big( f(w, Z) - f(w^*, Z) \big)
    \leq 12 \sqrt{ \frac{d}n} \cdot 3Lr \sqrt{\pi}
    \leq 64 Lr \sqrt{ \frac{d}n}.
\end{equation}
Since $|\eps (f(w, Z) - f(w^*, Z))| \leq B$ almost surely, we can apply Talagrand's concentration inequality for the supremum of the empirical process to
\[
    \sup\limits_{w \in \W[0,r]} P_n \eps \big( f(w, Z) - f(w^*, Z) \big).
\]
We use an upper bound on the exponential moment of the supremum, which follows from \cite[Theorems 2.1 and 2.3]{bousquet02}:
\begin{align}
    \label{eq:talagrand}
    &\notag
    \log \E_\eps \exp\left\{ \lambda \sup\limits_{w \in \W[0,r]} \sum\limits_{i=1}^n \eps_i \big( f(w, Z_i) - f(w^*, Z_i) \big) \right\}
    \\&
    \leq \lambda \E_\eps \sup\limits_{w \in \W[0,r]} \sum\limits_{i=1}^n \eps_i \big( f(w, Z_i) - f(w^*, Z_i) \big)
    + V_n \big(e^{B \lambda} - B \lambda - 1 \big)
    \quad \text{for all $\lambda \geq 0$},
\end{align}
where
\begin{align}
    \label{eq:variance_bound}
    V_n
    &\notag
    = \frac{2}{B} \E_\eps \sup\limits_{w \in \W[0,r]} \sum\limits_{i=1}^n \eps_i \big( f(w, Z) - f(w^*, Z) \big)
    + \frac{1}{B^2} \sup\limits_{w \in \W[0,r]} \sum\limits_{i=1}^n \big( f(w, Z) - f(w^*, Z) \big)^2
    \\&\qquad
    \leq \frac{2}{B} \E_\eps \sup\limits_{w \in \W[0,r]} \sum\limits_{i=1}^n \eps_i \big( f(w, Z) - f(w^*, Z) \big) + \frac{n L^2 r^2}{B^2}.
\end{align}
Combining the inequalities \eqref{eq:sup_expectation}, \eqref{eq:talagrand}, \eqref{eq:variance_bound} with the bound
$e^{x} - x - 1 \leq \frac{x^2 e^x}2,$ which holds for all $x \geq 0$,
we obtain for any $\lambda \geq 0$,
\begin{align*}
    &
    \log \E_\eps \exp\left\{ \lambda \sup\limits_{w \in \W[0,r]} P_n \eps \big( f(w, Z) - f(w^*, Z) \big) \right\}
    \\&
    \qquad\leq 64 \lambda Lr \sqrt{\frac{d}n} + \frac{B^2 \lambda^2 e^{B\lambda/n}}{2n} \left( \frac{128 Lr}{B} \sqrt{ \frac{d}n} + \frac{L^2 r^2}{B^2} \right).
\end{align*}
If $\lambda \leq n/B$, then
\[
    64 \lambda Lr \sqrt{\frac{d}n} + \frac{B^2 \lambda^2 e^{B\lambda/n}}{2n} \left( \frac{128 Lr}{B} \sqrt{ \frac{d}n} + \frac{L^2 r^2}{B^2} \right) \leq 64 (1 + e) \lambda Lr \sqrt{\frac{d}n} + \frac{e \lambda^2 L^2 r^2}{2n}.
\]
The claim follows.

\qed

\section{Conclusion}

In this work, we have established a tight high-probability upper bound on the excess risk of Empirical Risk Minimization (ERM), specifically $O(d / n + \log(1 / \delta) / n)$, for a subclass of exp-concave losses. The key aspect of our analysis is the convexity of the domain $\W$, though it does not necessitate boundedness. Nevertheless, it is important for the loss function $f(w, Z_i)$ to remain bounded for every $w \in \W$ and all $i \in \{1, \dots, n\}$. This assumption proves instrumental in Lemma \ref{lem:local_chaining}, when 
 we are applying Talagrand's inequality for the suprema of empirical processes. The counterpart to Talagrand's inequality in the unbounded case, as indicated in \citep{adamczak08}, might introduce an extra $\log n$ factor. Currently, it remains unclear whether a similar $O(d / n + \log(1 / \delta) / n)$ bound can be achieved in the unbounded scenario. While applications like logistic regression can mitigate the unbounded case to the bounded one with clipping (see \citep{foster2018logistic}), this results in nonconvex classes, making our localized analysis inapplicable. Moreover, as noted in Section \ref{sec:learning}, the additional $\log n$ factor is sometimes inavoidable when dealing with a non-convex reference set (see \citep[Theorem 6]{rakhlin2017empirical}). Understanding the geometry of non-convex classes, particularly in relation to proper and improper learners, appears to be a promising direction for future research.

\acks{The article was prepared within the framework of the HSE University Basic Research Program.  Nikita Puchkin is a Young Russian Mathematics award winner and would like to thank its sponsors and jury.}

\bibliography{references.bib}

\begin{thebibliography}{51}
\providecommand{\natexlab}[1]{#1}
\providecommand{\url}[1]{\texttt{#1}}
\expandafter\ifx\csname urlstyle\endcsname\relax
  \providecommand{\doi}[1]{doi: #1}\else
  \providecommand{\doi}{doi: \begingroup \urlstyle{rm}\Url}\fi

\bibitem[Adamczak(2008)]{adamczak08}
Radoslaw Adamczak.
\newblock {A tail inequality for suprema of unbounded empirical processes with
  applications to Markov chains}.
\newblock \emph{Electronic Journal of Probability}, 13:\penalty0 1000 -- 1034,
  2008.

\bibitem[Audibert and Catoni(2010)]{audibert2010linear}
Jean-Yves Audibert and Olivier Catoni.
\newblock Linear regression through {PAC}-{B}ayesian truncation.
\newblock \emph{arXiv preprint arXiv:1010.0072}, 2010.

\bibitem[Audibert and Catoni(2011)]{audibert2011robust}
Jean-Yves Audibert and Olivier Catoni.
\newblock Robust linear least squares regression.
\newblock \emph{The Annals of Statistics}, 39\penalty0 (5):\penalty0
  2766--2794, 2011.

\bibitem[Bach and Moulines(2013)]{bach2013non}
Francis Bach and Eric Moulines.
\newblock Non-strongly-convex smooth stochastic approximation with convergence
  rate ${O}(1/n)$.
\newblock \emph{Advances in Neural Information Processing Systems}, 26, 2013.

\bibitem[Bartlett et~al.(2005)Bartlett, Bousquet, and Mendelson]{bartlett05}
Peter~L. Bartlett, Olivier Bousquet, and Shahar Mendelson.
\newblock Local {R}ademacher complexities.
\newblock \emph{The Annals of Statistics}, 33\penalty0 (4):\penalty0
  1497--1537, 2005.

\bibitem[Bilodeau et~al.(2020)Bilodeau, Foster, and Roy]{bilodeau2020tight}
Blair Bilodeau, Dylan Foster, and Daniel Roy.
\newblock Tight bounds on minimax regret under logarithmic loss via
  self-concordance.
\newblock In \emph{International Conference on Machine Learning}, pages
  919--929, 2020.

\bibitem[Bilodeau et~al.(2021)Bilodeau, Foster, and Roy]{bilodeau2021minimax}
Blair Bilodeau, Dylan Foster, and Daniel Roy.
\newblock Minimax rates for conditional density estimation via empirical
  entropy.
\newblock \emph{arXiv preprint arXiv:2109.10461}, 2021.

\bibitem[Birg{\'e}(1983)]{birge1983approximation}
Lucien Birg{\'e}.
\newblock Approximation dans les espaces m{\'e}triques et th{\'e}orie de
  l'estimation.
\newblock \emph{Zeitschrift f{\"u}r Wahrscheinlichkeitstheorie und Verwandte
  Gebiete}, 65:\penalty0 181--237, 1983.

\bibitem[Birg{\'e} and Massart(1993)]{birge1993rates}
Lucien Birg{\'e} and Pascal Massart.
\newblock Rates of convergence for minimum contrast estimators.
\newblock \emph{Probability Theory and Related Fields}, 97:\penalty0 113--150,
  1993.

\bibitem[Bousquet(2002)]{bousquet02}
Olivier Bousquet.
\newblock A {B}ennett concentration inequality and its application to suprema
  of empirical processes.
\newblock \emph{Comptes Rendus Math\'{e}matique. Acad\'{e}mie des Sciences.
  Paris}, 334\penalty0 (6):\penalty0 495--500, 2002.

\bibitem[Bousquet et~al.(2020)Bousquet, Klochkov, and
  Zhivotovskiy]{bousquet2020sharper}
Olivier Bousquet, Yegor Klochkov, and Nikita Zhivotovskiy.
\newblock Sharper bounds for uniformly stable algorithms.
\newblock In \emph{Conference on Learning Theory}, pages 610--626, 2020.

\bibitem[Forster and Warmuth(2002)]{forster2002relative}
J{\"u}rgen Forster and Manfred~K Warmuth.
\newblock Relative expected instantaneous loss bounds.
\newblock \emph{Journal of Computer and System Sciences}, 64\penalty0
  (1):\penalty0 76--102, 2002.

\bibitem[Foster et~al.(2018)Foster, Kale, Luo, Mohri, and
  Sridharan]{foster2018logistic}
Dylan~J Foster, Satyen Kale, Haipeng Luo, Mehryar Mohri, and Karthik Sridharan.
\newblock Logistic regression: The importance of being improper.
\newblock In \emph{Conference On Learning Theory}, pages 167--208. PMLR, 2018.

\bibitem[Gaillard and Wintenberger(2018)]{gaillard18}
Pierre Gaillard and Olivier Wintenberger.
\newblock Efficient online algorithms for fast-rate regret bounds under
  sparsity.
\newblock In \emph{Advances in Neural Information Processing Systems},
  volume~31, 2018.

\bibitem[Gonen and Shalev-Shwartz(2018)]{gonen18}
Alon Gonen and Shai Shalev-Shwartz.
\newblock Average stability is invariant to data preconditioning.
  {I}mplications to exp-concave empirical risk minimization.
\newblock \emph{Journal of Machine Learning Research}, 18\penalty0
  (222):\penalty0 1--13, 2018.

\bibitem[Hazan et~al.(2014)Hazan, Koren, and Levy]{hazan14}
Elad Hazan, Tomer Koren, and Kfir~Y. Levy.
\newblock Logistic regression: Tight bounds for stochastic and online
  optimization.
\newblock In \emph{Proceedings of The 27th Conference on Learning Theory},
  volume~35, pages 197--209, 2014.

\bibitem[Kakade and Ng(2004)]{kakade2004online}
Sham~M Kakade and Andrew Ng.
\newblock Online bounds for bayesian algorithms.
\newblock \emph{Advances in Neural Information Processing Systems}, 17, 2004.

\bibitem[Kakade and Tewari(2008)]{kakade08}
Sham~M. Kakade and Ambuj Tewari.
\newblock On the generalization ability of online strongly convex programming
  algorithms.
\newblock In \emph{Advances in Neural Information Processing Systems},
  volume~21, 2008.

\bibitem[Kakade et~al.(2008)Kakade, Sridharan, and Tewari]{kakade08b}
Sham~M. Kakade, Karthik Sridharan, and Ambuj Tewari.
\newblock On the complexity of linear prediction: Risk bounds, margin bounds,
  and regularization.
\newblock In \emph{Advances in Neural Information Processing Systems},
  volume~21, 2008.

\bibitem[Kanade et~al.(2022)Kanade, Rebeschini, and
  Va{\v{s}}kevi{\v{c}}ius]{kanade2022exponential}
Varun Kanade, Patrick Rebeschini, and Tomas Va{\v{s}}kevi{\v{c}}ius.
\newblock Exponential tail local {R}ademacher complexity risk bounds without
  the {B}ernstein condition.
\newblock \emph{arXiv preprint arXiv:2202.11461}, 2022.

\bibitem[Klochkov and Zhivotovskiy(2021)]{klochkov21}
Yegor Klochkov and Nikita Zhivotovskiy.
\newblock Stability and deviation optimal risk bounds with convergence rate
  ${O}(1/n)$.
\newblock In \emph{Advances in Neural Information Processing Systems},
  volume~34, pages 5065--5076, 2021.

\bibitem[Koltchinskii(2006)]{koltchinskii2006local}
Vladimir Koltchinskii.
\newblock Local {R}ademacher complexities and oracle inequalities in risk
  minimization.
\newblock \emph{The Annals of Statistics}, 34\penalty0 (6):\penalty0 2593 --
  2656, 2006.

\bibitem[Koltchinskii(2011)]{koltchinskii2011oracle}
Vladimir Koltchinskii.
\newblock \emph{Oracle Inequalities in Empirical Risk Minimization and Sparse
  Recovery Problems: Ecole d'Et{\'e} de Probabilit{\'e}s de Saint-Flour
  XXXVIII-2008}, volume 2033.
\newblock Springer Science \& Business Media, 2011.

\bibitem[Koren and Levy(2015)]{koren15}
Tomer Koren and Kfir Levy.
\newblock Fast rates for exp-concave empirical risk minimization.
\newblock In \emph{Advances in Neural Information Processing Systems},
  volume~28, 2015.

\bibitem[LeCam(1973)]{lecam1973convergence}
Lucien LeCam.
\newblock Convergence of estimates under dimensionality restrictions.
\newblock \emph{The Annals of Statistics}, pages 38--53, 1973.

\bibitem[Lecu{\'e} and Rigollet(2014)]{lecue2014optimal}
Guillaume Lecu{\'e} and Philippe Rigollet.
\newblock {Optimal learning with $Q$-aggregation}.
\newblock \emph{The Annals of Statistics}, 42\penalty0 (1):\penalty0 211 --
  224, 2014.

\bibitem[Liang et~al.(2015)Liang, Rakhlin, and Sridharan]{liang15}
Tengyuan Liang, Alexander Rakhlin, and Karthik Sridharan.
\newblock Learning with square loss: Localization through offset {R}ademacher
  complexity.
\newblock In \emph{Proceedings of The 28th Conference on Learning Theory},
  volume~40, pages 1260--1285, 2015.

\bibitem[Mahdavi et~al.(2015)Mahdavi, Zhang, and Jin]{mahdavi2015lower}
Mehrdad Mahdavi, Lijun Zhang, and Rong Jin.
\newblock Lower and upper bounds on the generalization of stochastic
  exponentially concave optimization.
\newblock In \emph{Conference on Learning Theory}, pages 1305--1320, 2015.

\bibitem[Massart and N\'{e}d\'{e}lec(2006)]{massart06}
Pascal Massart and \'{E}lodie N\'{e}d\'{e}lec.
\newblock Risk bounds for statistical learning.
\newblock \emph{The Annals of Statistics}, 34\penalty0 (5):\penalty0
  2326--2366, 2006.

\bibitem[Mehta(2017)]{mehta17}
Nishant Mehta.
\newblock {Fast rates with high probability in exp-concave statistical
  learning}.
\newblock In \emph{Proceedings of the 20th International Conference on
  Artificial Intelligence and Statistics}, volume~54, pages 1085--1093, 2017.

\bibitem[Mendelson(2002)]{mendelson2002geometric}
Shahar Mendelson.
\newblock Geometric parameters of kernel machines.
\newblock In \emph{International Conference on Computational Learning Theory},
  pages 29--43. Springer, 2002.

\bibitem[Moulines and Bach(2011)]{moulines11}
Eric Moulines and Francis Bach.
\newblock Non-asymptotic analysis of stochastic approximation algorithms for
  machine learning.
\newblock In \emph{Advances in Neural Information Processing Systems},
  volume~24, 2011.

\bibitem[Mourtada and Ga\"{i}ffas(2022)]{mourtada2022improper}
Jaouad Mourtada and St\'{e}phane Ga\"{i}ffas.
\newblock An improper estimator with optimal excess risk in misspecified
  density estimation and logistic regression.
\newblock \emph{Journal of Machine Learning Research}, 23\penalty0
  (31):\penalty0 1--49, 2022.

\bibitem[Mourtada et~al.(2021)Mourtada, Va{\v{s}}kevi{\v{c}}ius, and
  Zhivotovskiy]{Mourtada2021}
Jaouad Mourtada, Tomas Va{\v{s}}kevi{\v{c}}ius, and Nikita Zhivotovskiy.
\newblock Distribution-free robust linear regression.
\newblock \emph{Mathematical Statistics and Learning}, 4\penalty0
  (3-4):\penalty0 253--292, 2021.

\bibitem[Polyanskiy and Wu(2023)]{polyanskiy2023}
Yury Polyanskiy and Yihong Wu.
\newblock \emph{Information Theory: From Coding to Learning}.
\newblock Cambridge University Press, 2023.

\bibitem[Rakhlin and Sridharan(2015)]{rakhlin2015sequential}
Alexander Rakhlin and Karthik Sridharan.
\newblock Sequential probability assignment with binary alphabets and large
  classes of experts.
\newblock \emph{arXiv preprint arXiv:1501.07340}, 2015.

\bibitem[Rakhlin et~al.(2017)Rakhlin, Sridharan, and
  Tsybakov]{rakhlin2017empirical}
Alexander Rakhlin, Karthik Sridharan, and Alexandre~B. Tsybakov.
\newblock Empirical entropy, minimax regret and minimax risk.
\newblock \emph{Bernoulli}, 23\penalty0 (2):\penalty0 789--824, 2017.
\newblock ISSN 1350-7265.

\bibitem[Shamir(2015)]{shamir15}
Ohad Shamir.
\newblock The sample complexity of learning linear predictors with the squared
  loss.
\newblock \emph{Journal of Machine Learning Research}, 16\penalty0
  (108):\penalty0 3475--3486, 2015.

\bibitem[Sridharan et~al.(2008)Sridharan, Shalev-Shwartz, and
  Srebro]{sridharan08}
Karthik Sridharan, Shai Shalev-Shwartz, and Nathan Srebro.
\newblock Fast rates for regularized objectives.
\newblock In \emph{Advances in Neural Information Processing Systems},
  volume~21, 2008.

\bibitem[Van~de Geer(1990)]{van1990estimating}
Sara Van~de Geer.
\newblock Estimating a regression function.
\newblock \emph{The Annals of Statistics}, pages 907--924, 1990.

\bibitem[Van~de Geer(2000)]{van2000empirical}
Sara Van~de Geer.
\newblock \emph{Empirical Processes in M-estimation}.
\newblock Cambridge University Press, 2000.

\bibitem[Van~der Hoeven et~al.(2022)Van~der Hoeven, Zhivotovskiy, and
  Cesa-Bianchi]{van2022regret}
Dirk Van~der Hoeven, Nikita Zhivotovskiy, and Nicol{\`o} Cesa-Bianchi.
\newblock A regret-variance trade-off in online learning.
\newblock \emph{arXiv preprint arXiv:2206.02656}, 2022.

\bibitem[Va{\v{s}}kevi{\v{c}}ius and
  Zhivotovskiy(2023)]{vavskevivcius2023suboptimality}
Tomas Va{\v{s}}kevi{\v{c}}ius and Nikita Zhivotovskiy.
\newblock Suboptimality of constrained least squares and improvements via
  non-linear predictors.
\newblock \emph{Bernoulli}, 29\penalty0 (1):\penalty0 473--495, 2023.

\bibitem[Vijaykumar(2021)]{vijaykumar21}
Suhas Vijaykumar.
\newblock Localization, convexity, and star aggregation.
\newblock In \emph{Advances in Neural Information Processing Systems},
  volume~34, pages 4570--4581, 2021.

\bibitem[Wegkamp(2003)]{wegkamp2003model}
Marten Wegkamp.
\newblock Model selection in nonparametric regression.
\newblock \emph{The Annals of Statistics}, 31\penalty0 (1):\penalty0 252--273,
  2003.

\bibitem[Woodworth and Srebro(2021)]{woodworth21}
Blake~E. Woodworth and Nathan Srebro.
\newblock An even more optimal stochastic optimization algorithm: Minibatching
  and interpolation learning.
\newblock In \emph{Advances in Neural Information Processing Systems},
  volume~34, pages 7333--7345, 2021.

\bibitem[Yang et~al.(2018)Yang, Li, and Zhang]{yang2018simple}
Tianbao Yang, Zhe Li, and Lijun Zhang.
\newblock A simple analysis for exp-concave empirical minimization with
  arbitrary convex regularizer.
\newblock In \emph{International Conference on Artificial Intelligence and
  Statistics}, pages 445--453, 2018.

\bibitem[Yang and Barron(1999)]{yang1999information}
Yuhong Yang and Andrew Barron.
\newblock Information-theoretic determination of minimax rates of convergence.
\newblock \emph{Annals of Statistics}, pages 1564--1599, 1999.

\bibitem[Zhang and Zhou(2019)]{zhang19}
Lijun Zhang and Zhi-Hua Zhou.
\newblock Stochastic approximation of smooth and strongly convex functions:
  Beyond the ${O}(1/t)$ convergence rate.
\newblock In \emph{Proceedings of the Thirty-Second Conference on Learning
  Theory}, volume~99, pages 3160--3179, 2019.

\bibitem[Zhang(2006)]{zhang2006information}
Tong Zhang.
\newblock Information-theoretic upper and lower bounds for statistical
  estimation.
\newblock \emph{IEEE Transactions on Information Theory}, 52\penalty0
  (4):\penalty0 1307--1321, 2006.

\bibitem[Zhivotovskiy and Hanneke(2018)]{zhivotovskiy18}
Nikita Zhivotovskiy and Steve Hanneke.
\newblock Localization of {VC} classes: Beyond local {R}ademacher complexities.
\newblock \emph{Theoretical Computer Science}, 742:\penalty0 27--49, 2018.

\end{thebibliography}

\appendix

\section{Proofs of auxiliary results}
\label{app:propositions}

\subsection{Proof of Proposition \ref{prof:exp-concavity}}

It is enough to check that
\[
    \nabla^2 e^{-\sigma \widehat F(w) / L^2} \preceq \mathsf{O}
\]
for all $w$ from the interior of $\W$. Here $\mathsf{O}$ is a matrix with zero entries. This is equivalent to
\[
    \frac{\sigma}{L^2} \nabla \widehat F(w) \nabla \widehat F(w)^\top \preceq \nabla^2 \widehat F(w)
    \quad \text{for all $w \in \text{Int}(\W)$}.
\]
The strong convexity of $\widehat F$ with respect to the seminorm $\|\cdot\|_\H$ implies that the function $\widehat F(w) - \sigma \|w\|_\H^2$ is convex on $\H$. The last is equivalent to the inequality
\[
    \nabla^2 \left( \widehat F(w) - \sigma \|w\|_\H^2 \right) \succeq \mathsf{O}
    \quad \text{for all $w \in \text{Int}(\W)$}
\]
or simply $\nabla^2 \widehat F(w) \succeq \sigma \H$ for all $w \in \text{Int}(\W)$. On the other hand, it holds that
\[
    \big( \widehat F(u) - \widehat F(v) \big)^2
    \leq P_n \big( f(u, Z) - f(v, Z) \big)^2
    \leq L^2 \|u - v\|_\H^2
    \quad \text{for all $u, v \in \W$.}
\]
This yields that
\[
    \nabla \widehat F(w) \nabla \widehat F(w)^\top \preceq L^2 \H \quad \text{for all $w \in \text{Int}(\W)$}.
\]
Hence, for any $w$ in the interior of $\W$, we have
\[
    \frac\sigma{L^2} \nabla \widehat F(w) \nabla \widehat F(w)^\top
    \preceq \sigma \H
    \preceq \nabla^2 \widehat F(w).
\]
The claim follows.

\hfill\qed

\subsection{Proof of Proposition \ref{prop:boundedness}}

If Assumption \ref{as:sc} holds, then Jensen's inequality immediately implies that
\[
    \big( \widehat F(u) - \widehat F(v) \big)^2
    \leq P_n \big( f(u, Z) - f(v, Z) \big)^2
    \leq L^2 \|u - v\|_\H^2
    \quad \text{for all $u, v \in \W$.}
\]
Thus, $\widehat F$ is $L$-Lipschitz on $\W$ with respect to the seminorm $\|\cdot\|_\H$.
On the other hand, the strong convexity of $\widehat F$ yields that
\[
    \widehat F(\widehat w)
    \leq \widehat F\left(\frac{\widehat w + w}2 \right) \leq \frac12 \widehat F(\widehat w) + \frac12 \widehat F(w) - \frac{\sigma}8 \|\widehat w - w\|_\H^2
    \quad \text{for all $w \in \W$,}
\]
where the first inequality follows from the convexity of $\W$ and the definition of $\widehat w$. Hence,
\[
    \widehat F(w) \geq \widehat F(\widehat w) + \frac{\sigma}4 \|\widehat w - w\|_\H^2
    \quad \text{for all $w \in \W$.}
\]
As a result, we obtain that
\[
    \left( \widehat F(w) - \widehat F(\widehat w) \right)^2 \leq L^2 \|\widehat w - w\|_\H^2
    \leq \frac{4 L^2}{\sigma} \left( \widehat F(w) - \widehat F(\widehat w) \right)
    \quad \text{for all $w \in \W$.}
\]
Consequently,
\begin{equation}
    \label{eq:hat_f_bound}
    \left| \widehat F(w) - \widehat F(w^*) \right|
    \leq \max\left\{ \widehat F(w) - \widehat F(\widehat w), \widehat F(w^*) - \widehat F(\widehat w) \right\}
    \leq \frac{4 L^2}{\sigma}
    \quad \text{for all $w \in \W$.}
\end{equation}
We show that $|f(w, Z) - f(w^*, Z)| \leq 4 L^2 / \sigma$ almost surely. Assume the opposite. Let $\Z_> \subseteq \Z$ be a set of all such $z \in \Z$ that
$f(w, z) - f(w^*, z) > 4 L^2 / \sigma$. If $\Z_>$ has a positive probability measure, then we have $\widehat F(w) - \widehat F(w^*) > 4 L^2 / \sigma$ on $\Z_>^{\otimes n}$, which contradicts \eqref{eq:hat_f_bound}. Hence, $f(w, Z) - f(w^*, Z) \leq 4 L^2 / \sigma$ almost surely. Similarly, we can prove that $f(w^*, Z) - f(w, Z) \leq 4 L^2 / \sigma$ almost surely.

\hfill\qed

\subsection{Proof of Lemma \ref{lem:offset_symmetrization}} 
\label{app:offset_symmetrization_proof}

Note that
\begin{align*}
    F(\widehat w) - F(w^*)
    &
    = 2 (P - P_n) \big( f(\widehat w, Z) - f(w^*, Z) \big)
    + P \big( f(w^*, Z) - f(\widehat w, Z) \big)
    \\&\quad
    + 2 P_n \big( f(\widehat w, Z) - f(w^*, Z) \big).
\end{align*}
According to Assumption \ref{as:sc}, $\widehat F$ is strongly convex with respect to the seminorm, induced by the matrix $\H$. 
Taking into account the convexity of $\W$ and the equation \eqref{eq:sc}, we obtain that
\[
    \widehat F(\widehat w)
    \leq \widehat F\left(\frac{\widehat w + w}2 \right) \leq \frac12 \widehat F(\widehat w) + \frac12 \widehat F(w) - \frac{\sigma}8 \|\widehat w - w\|_\H^2
    \quad \text{for all $w \in \W$.}
\]
Therefore, it holds that
\begin{equation}
    \label{eq:sc_opt1}
    \widehat F(w) \geq \widehat F(\widehat w) + \frac{\sigma}4 \|\widehat w - w\|_\H^2
    \quad \text{for all $w \in \W$.}
\end{equation}
Similarly, we deduce that the following inequality holds for all $w \in \W$:
\[
    F(w^*)
    = \E \widehat F(w^*)
    \leq \E \widehat F\left(\frac{w^* + w}2 \right)
    \leq \frac12 \E \widehat F(w^*) + \frac12 \E \widehat F(w) - \frac{\sigma}8 \E \|w^* - w\|_\H^2.
\]
Consequently, we have
\begin{equation}
    \label{eq:sc_opt2}
    F(w) \geq F(w^*) + \frac{\sigma}4 \E \|w^* - w\|_{\H}^2
    \quad \text{for all $w \in \W$.}
\end{equation}
Observe that $\E\|w^* - w\|_{\H}^2 = \|w^* - w\|_{\E\H}^2$.
Then the inequalities \eqref{eq:sc_opt1} and \eqref{eq:sc_opt2} imply that
\[
    P\big( f(w^*, Z) - f(\widehat w, Z) \big)
    \leq - \frac{\sigma}4 \|\widehat w - w^*\|_{\E\H}^2
    \quad \textrm{and} \quad P_n \big( f(\widehat w, Z) - f(w^*, Z) \big)
    \leq - \frac{\sigma}4 \|\widehat w - w^*\|_{\H}^2.
\]
Thus, it holds that
\[
    F(\widehat w) - F(w^*)
    \leq 2 (P - P_n) \big( f(\widehat w, Z) - f(w^*, Z) \big)
    - \frac{\sigma}4 \|\widehat w - w^*\|_{\E \H}^2
    - \frac{\sigma}2 \|\widehat w - w^*\|_{\H}^2. 
\]   
The last expression does not exceed
\[
    2 \sup\limits_{w \in \W} \left[ (P - P_n) \big( f(w, Z) - f(w^*, Z) \big) - \frac{\sigma}8 \left( \|w - w^*\|_{\E \H}^2 + \|w - w^*\|_{\H}^2 \right) \right].
\]
We are ready to apply the symmetrization argument. Let $Z_1', \dots, Z_n'$ be independent copies of $Z_1, \dots, Z_n$, and let $\H' = \H(Z'_1, \dots, Z'_n)$ be the corresponding matrix from Assumption \ref{as:sc}.
Then
\begin{align*}
    &
    \E \Phi \left( 2 \sup\limits_{w \in \W} \left[ (P - P_n) \big( f(w, Z) - f(w^*, Z) \big) - \frac{\sigma}8 \left( \|w - w^*\|_{\E \H}^2 + \|w - w^*\|_{\H}^2 \right) \right] \right)
    \\&
    = \E \Phi \left(  2 \sup\limits_{w \in \W} \left[ \E' (P_n' - P_n) \big( f(w, Z) - f(w^*, Z) \big) - \frac{\sigma}8 \left(\E' \|w - w^*\|_{\H'}^2 + \|w - w^*\|_{\H}^2 \right) \right] \right),
\end{align*}
where $P_n'$ denotes the expectation with respect to the uniform measure on $Z_1', \dots, Z_n'$ and $\E'$ stands for the expectation with respect to $Z_1', \dots, Z_n'$. Using Jensen's inequality, we obtain that
\begin{align*}
    &
    \E \Phi \left( 2 \sup\limits_{w \in \W} \left[ (P - P_n) \big( f(w, Z) - f(w^*, Z) \big) - \frac{\sigma}8 \left( \|w - w^*\|_{\E \H}^2 + \|w - w^*\|_{\H}^2 \right) \right] \right)
    \\&
    \leq \E \E' \Phi \left(  2 \sup\limits_{w \in \W} \left[ (P_n' - P_n) \big( f(w, Z) - f(w^*, Z) \big) - \frac{\sigma}8 \left(\|w - w^*\|_{\H'}^2 + \|w - w^*\|_{\H}^2 \right) \right] \right).
\end{align*}
Let $\eps_1, \dots, \eps_n$ be i.i.d. Rademacher random variables. Note that, whatever the realization of $\eps_1, \dots, \eps_n$ is,
\[
    (P_n' - P_n) \eps \big( f(w, Z) - f(w^*, Z) \big) - \frac{\sigma}8 \left(\|w - w^*\|_{\H'}^2 + \|w - w^*\|_{\H}^2 \right)
\]
has the same distribution as
\[
    (P_n' - P_n) \big( f(w, Z) - f(w^*, Z) \big) - \frac{\sigma}8 \left(\|w - w^*\|_{\H'}^2 + \|w - w^*\|_{\H}^2 \right).
\]
Thus,
\begin{align*}
    &
    \E \E' \Phi \left( 2 \sup\limits_{w \in \W} \left[ (P_n' - P_n) \big( f(w, Z) - f(w^*, Z) \big) - \frac{\sigma}8 \left(\|w - w^*\|_{\H'}^2 + \|w - w^*\|_{\H}^2 \right) \right] \right)
    \\&
    = \E \E' \E_\eps \Phi \left( 2 \sup\limits_{w \in \W} \left[ (P_n' - P_n) \eps \big( f(w, Z) - f(w^*, Z) \big) - \frac{\sigma}8 \left(\|w - w^*\|_{\H'}^2 + \|w - w^*\|_{\H}^2 \right) \right] \right)
    \\&
    \leq \E \E_\eps \Phi \left( 4 \sup\limits_{w \in \W} \left[ P_n \eps \big( f(w, Z) - f(w^*, Z) \big) - \frac{\sigma}8 \|w - w^*\|_\H^2 \right] \right),
\end{align*}
where the last line is due to the triangle inequality. Hence, we conclude that
\begin{align*}
    &
    \E \Phi \big( F(\widehat w) - F(w^*) \big)
    \\&
    \leq \E \E_\eps \Phi \left(2 \sup\limits_{w \in \W} \left[ (P - P_n) \big( f(w, Z) - f(w^*, Z) \big) - \frac{\sigma}8 \left( \|w - w^*\|_{\E \H}^2 + \|w - w^*\|_{\H}^2 \right) \right] \right)
    \\&
    \leq \E \E_\eps \Phi \left(4 \sup\limits_{w \in \W} \left[ P_n \eps \big( f(w, Z) - f(w^*, Z) \big) - \frac{\sigma}8 \|w - w^*\|_\H^2 \right] \right).
\end{align*}
The claim follows.

\qed

\subsection{Proof of Lemma \ref{lem:covering}}
\label{sec:covering_proof}

Let $
    \H = \sum\limits_{j=1}^{\text{rank}(\H)} \lambda_j v_j v_j^\top
$
be the eigendecomposition of $\H$. Here we assume that $\lambda_1 \geq \lambda_2 \geq \dots \geq \lambda_{\text{rank}(\H)} > 0$, and the vectors $v_1, \dots, v_{\text{rank}(\H)}$ form an orthonormal system.
Define the Moore-Penrose pseudoinverse of $\H$
\[
    \H^\dag = \sum\limits_{j=1}^{\text{rank}(\H)} \lambda_j^{-1} v_j v_j^\top
\]
and the square roots
\[
    \H^{1/2} = \sum\limits_{j=1}^{\text{rank}(\H)} \sqrt{\lambda_j} v_j v_j^\top \quad \textrm{and} \quad \left(\H^\dag \right)^{1/2} = \sum\limits_{j=1}^{\text{rank}(\H)} \frac1{\sqrt{\lambda_j}} v_j v_j^\top.
\]
It is easy to observe that
\[
    \Pi = \H^{1/2} \left(\H^\dag \right)^{1/2} = \sum\limits_{j=1}^{\text{rank}(\H)} v_j v_j^\top.
\]
is nothing but the orthogonal projector onto the image of $\H^{1/2}$.
Let $\{a_1, \dots, a_N\}$ be the smallest $u$-net of the Euclidean ball $\B(0, r) \subset \R^d$ with respect to the Euclidean norm. It is known that $N \leq (6r/u)^d$. We show that $\left\{w^* + \left(\H^\dag \right)^{1/2} a_1, \dots, w^* + \left(\H^\dag \right)^{1/2} a_N\right\}$ is a $u$-net of $\W[0, r]$ with respect to the seminorm $\|\cdot\|_\H$.
For this purpose, fix any $w \in \W[0,r]$ and let
\[
    a = \H^{1/2}(w - w^*).
\]
By the definition of $\Pi$ and $a$, $\Pi a = a$.
Find the closest to $a$ element amongst $\{a_1, \dots, a_N\}$ (with respect to the Euclidean distance). Without loss of generality, assume that it is $a_1$.
Denote $w^* + \left(\H^\dag \right)^{1/2} a_1$ by $w_1$. Then, it holds that
\begin{align*}
    \|w - w_1\|_\H
    &
    = \|\H^{1/2}(w - w_1)\|
    = \|a - \H^{1/2}(w_1 - w^*)\|
    \\&
    = \left\|a - \H^{1/2}\left(\H^\dag \right)^{1/2} a_1 \right\|
    = \|a - \Pi a_1\|
    = \|\Pi a - \Pi a_1\|
    \leq \|a - a_1\| \leq u.
\end{align*}
Hence, $\left\{w^* + \left(\H^\dag \right)^{1/2} a_1, \dots, w^* + \left(\H^\dag \right)^{1/2} a_N\right\}$ is a $u$-net of $\W[0, r]$ with respect to $\|\cdot\|_\H$. The claim follows.

\hfill\qed

\end{document}